\documentclass{article}

\usepackage{times}
\usepackage{graphicx} 
\usepackage{subfigure} 

\usepackage{natbib}

\usepackage{algorithm}
\usepackage{algorithmic}

\usepackage{hyperref}

\usepackage[accepted]{icml2014} 

\usepackage{amssymb,amsfonts,amsbsy,amsmath,amsthm}
\theoremstyle{plain}
\newtheorem{thm}{Theorem}[section]
\newtheorem{theorem}[thm]{Theorem}

\newtheorem{prop}[thm]{Proposition}
\newtheorem{proposition}[thm]{Proposition}

\theoremstyle{definition}

\newtheorem{example}[thm]{Example}

\newtheorem{definition}[thm]{Definition}

\DeclareMathOperator*{\argmax}{\arg\!\max}

\icmltitlerunning{Exchangeable Variable Models}

\begin{document} 

\twocolumn[
\icmltitle{Exchangeable Variable Models}

\icmlauthor{Mathias Niepert}{mniepert@cs.washington.edu}
\icmlauthor{Pedro Domingos}{pedrod@cs.washington.edu}
\icmladdress{Department of Computer Science \& Engineering,
            University of Washington, Seattle, WA 98195, USA}

\icmlkeywords{exchangeable variables, machine learning, classification}

\vskip 0.3in
]

\begin{abstract} 
A sequence of random variables is exchangeable if its joint distribution is invariant under variable permutations. We introduce exchangeable variable models (EVMs) as a novel class of probabilistic models whose basic building blocks are partially exchangeable sequences, a generalization of exchangeable sequences. We prove that a family of tractable EVMs is optimal under zero-one loss for a large class of functions, including parity and threshold functions, and strictly subsumes existing tractable independence-based model families. Extensive experiments show that EVMs outperform state of the art classifiers such as SVMs and  probabilistic models which are solely based on independence assumptions.
\end{abstract} 

\section{Introduction}
\label{intro}

\begin{figure}[h!]
\begin{center}
\includegraphics[width=0.475\textwidth]{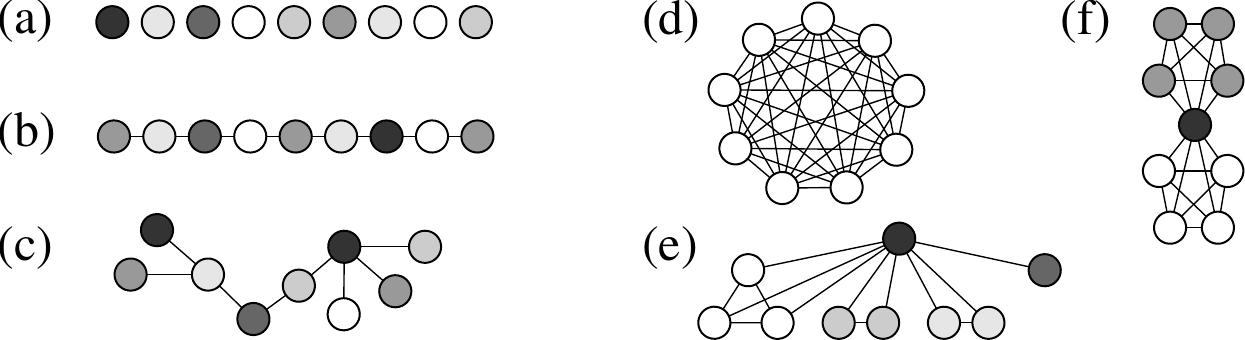}
\caption{\label{fig:sym} Illustration of low tree-width models exploiting independence (a)-(c) and exchangeable variable models (EVMs) exploiting finite exchangeability (variable nodes with identical shades are exchangeable) (d)-(f).}
\end{center}
\end{figure}

Conditional independence is a crucial notion that facilitates efficient inference and parameter learning in probabilistic models. Its logical and algorithmic properties as well as its graphical representations have led to the advent of graphical models as a discipline within artificial intelligence~\cite{Koller:2009}. The notion of finite (partial) exchangeability~\cite{Diaconis:1980}, on the other hand, has not yet been explored as a basic building block for tractable probabilistic models. A sequence of random variables is exchangeable if its distribution is invariant under variable permutations. Similar to conditional independence, partial exchangeability, a generalization of exchangeability, can  reduce the complexity of parameter learning and is a concept that facilitates high tree-width graphical models with tractable inference. For instance, the graphical models (a)-(c) with Bernoulli variables in Figure~\ref{fig:sym} depict typical low tree-width models based on the notion of (conditional) independence. Graphical models (d)-(f) have high tree-width but are tractable if we assume the variables with identical shades to be exchangeable. 
We will see that EVMs are especially beneficial for high-dimensional and sparse domains such as text and collaborative filtering problems.
While there exists work on tractable models, with a majority focusing on low tree-width graphical models, a framework for finite partial exchangeability as a basic building block of \emph{tractable} probabilistic models seems natural but does not yet exist. 

We propose exchangeable variable models (EVMs), a novel family of probabilistic models for classification and probability estimation. While most probabilistic models are built on the notion of conditional independence and its graphical representation, EVMs have finite partially exchangeable sequences as basic components. We show that EVMs can represent complex positive and negative correlations between large sets of variables with few parameters and without sacrificing tractable inference. 
The parameters of EVMs are estimated under the maximum-likelihood principle and we assume the examples to be independent and identically distributed. 
We develop methods for efficient probabilistic inference, maximum-likelihood estimation, and structure learning.

We introduce the mixtures of EVMs (MEVMs) family of models which is strictly more expressive than the naive Bayes family of models but as efficient to learn. MEVMs represent classifiers that are optimal under zero-one loss for a large class of Boolean functions including parity and threshold functions. Extensive experiments show that exchangeable variable models, when combined with the notion of conditional independence, are effective both for classification and probability estimation. The MEVM classifier significantly outperforms  state of the art classifiers on numerous high-dimensional and sparse data sets. MEVMs also outperform several tractable graphical model classes on typical probability estimation problems while being orders of magnitudes more efficient.

\section{Background}

We begin by reviewing the statistical concepts of finite exchangeability and finite partial exchangeability.

\subsection{Finite Exchangeability}

Finite exchangeability is best understood in the context of a finite sequence of binary random variables such as a finite number of coin tosses. Here, finite exchangeability means that it is only the number of heads that matters and not their particular order. Since exchangeable variables are not necessarily independent, finite exchangeability can model highly correlated variables, a graphical representation of which would be the fully connected graph with high tree-width (see Figure~\ref{fig:sym}(d)). However, as we will later see, the number of parameters and the complexity of inference remains linear in the number of variables.

\begin{definition}[Exchangeability]
\label{full-exch}
Let $X_1, ...,X_n$ be a sequence of random variables with joint distribution $P$ and let $S(n)$ be the group of all permutations acting on $\{1, ..., n\}$. We say that $X_1, ...,X_n$ is exchangeable if $P(X_1, ...,X_n) = P(X_{\pi(1)}, ... ,X_{\pi(n)})$
for all $\pi \in S(n)$.
\end{definition}

In this paper, we are concerned with exchangeable \emph{variables} and iid \emph{examples}. The literature has mostly focused on exchangeability of an \emph{infinite} sequence of random  variables. In this case, one can express the joint distribution as a mixture of iid sequences~\cite{finetti:1938}. However, for finite sequences of exchangeable variables this representation is inadequate -- while finite exchangeable sequences can be approximated with de Finetti style mixtures of iid sequences, these approximations are not suitable for finite sequences of random variables not extendable to an infinite exchangeable sequence~\cite{DnF:1980}. Moreover, negative correlations can only be modeled in the finite case. There are interesting connections between the automorphisms of graphical models and finite exchangeability~\cite{niepertorbits}. An alternative approach to exchangeability considers its relationship to sufficiency~\cite{Diaconis:1980,lauritzen:1984} which is at the core of our work.

\subsection{Finite Partial Exchangeability}

The assumption that all variables of a probabilistic model are exchangeable is often too strong. Fortunately, finite exchangeability can be generalized to the concept of finite partial exchangeability using the notion of a statistic. 

\begin{definition}[Partial Exchangeability]
Let $X_1, ...,X_n$ be a sequence of random variables with distribution $P$, let  $\mathbf{Val}(X_i)$ be the domain of $X_i$, and let $\mathcal{T}$ be a finite set. The sequence $X_1, ...,X_n$ is partially exchangeable with respect to the statistic $T: \mathbf{Val}(X_1)\times ...\times \mathbf{Val}(X_n)\rightarrow \mathcal{T}$ if
$$T(\mathbf{x})=T(\mathbf{x'}) \mbox { implies } P(\mathbf{x})=P(\mathbf{x'}),$$ where $\mathbf{x}$ and $\mathbf{x'}$ are assignments to the sequence of random variables $X_1,...,X_n$.
\end{definition}

The following theorem states that the joint distribution of a sequence of random variables, which is partially exchangeable with respect to a statistic $T$, is a unique mixture of uniform distributions.

\begin{theorem}\emph{\cite{Diaconis:1980}}
\label{theorem-param-pe}
Let $X_1, ...,X_n$ be a sequence of random variables with distribution $P$, let $\mathcal{T}$ be a finite set, and let $T: \mathbf{Val}(X_1)\times ...\times \mathbf{Val}(X_n)\rightarrow \mathcal{T}$ be a statistic. Moreover, let $S_t = \{\mathbf{x} \in \mathbf{Val}(X_1)\times ...\times \mathbf{Val}(X_n) \mid T(\mathbf{x})=t\}$, let $U_t$ be the uniform distribution on $S_t$, and let $w_t = P(S_t)$. If $X_1, ...,X_n$ is partially exchangeable with respect to $T$, then
\begin{equation}
\label{equation-paramterization}
P(\mathbf{x}) = \sum_{t \in \mathcal{T}} w_t U_t(\mathbf{x}).
\end{equation}
\end{theorem}

\begin{figure}[t!]
\begin{center}
\includegraphics[width=0.37\textwidth]{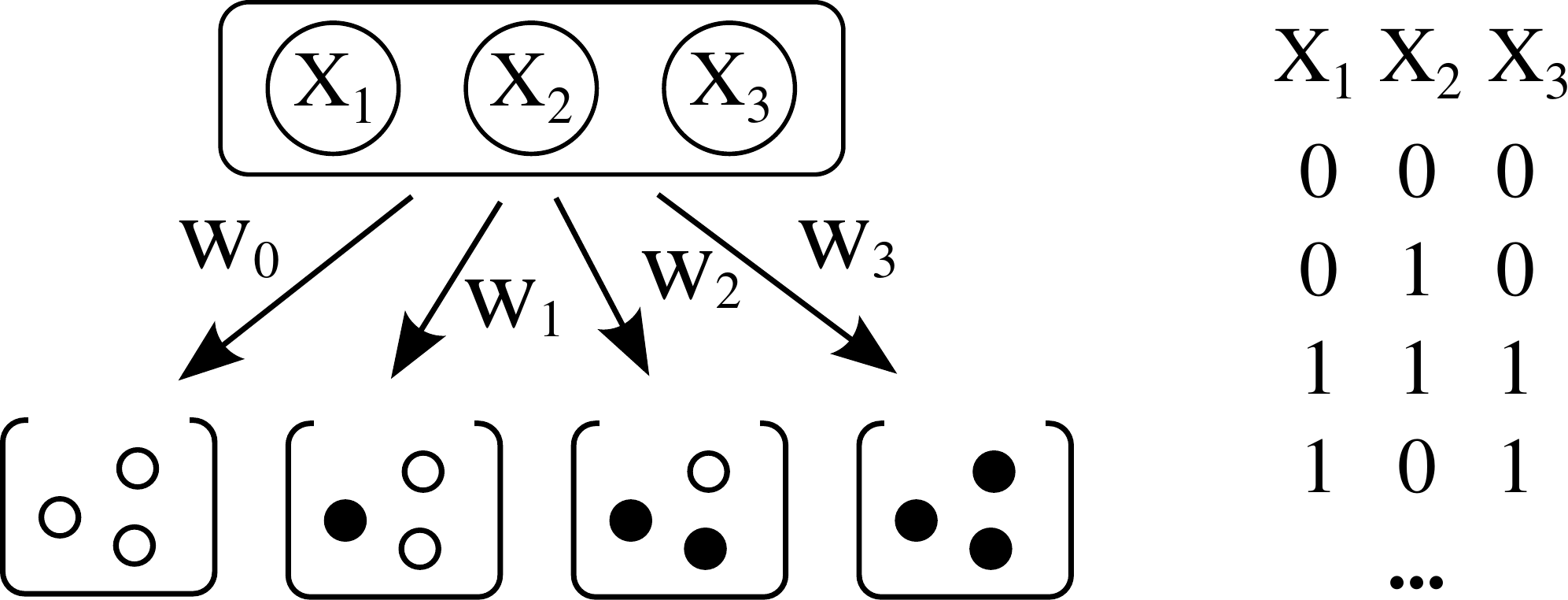}
\caption{\label{fig:urn} A finite sequence of exchangeable variables can be parameterized as a unique mixture of urn processes. Each such urn process is a series of draws without replacement. }
\end{center}
\end{figure}

The theorem provides an implicit description of the distributions $U_t$. The challenge for specific families of random variables lies in finding a statistic $T$ with respect to which a sequence of variables is partially exchangeable and an efficient algorithm to compute the probabilities $U_t(\mathbf{x})$. For the case of exchangeable sequences of discrete random variables and, in particular, exchangeable sequences of binary random variables, an explicit description does exist and is well-known in the statistics literature \cite{Diaconis:1980,Stefanescu:2003}.

\begin{example}
\label{example-exch}
Let $X_1,X_2,X_3$ be three exchangeable binary variables with joint distribution $P$. Then, the sequence $X_1,X_2,X_3$ is partially exchangeable with respect to the statistic $T: \{0,1\}^3 \rightarrow \mathcal{T} = \{0,1,2,3\}$ with $T(\mathbf{x}=(x_1,x_2,x_3)) = x_1+x_2+x_3.$ Thus, we can write 
$$P(\mathbf{x}) = \sum_{t \in \mathcal{T}} w_t U_t(\mathbf{x}),$$ where $w_t = P(T(\mathbf{x}) = t)$, $U_t(\mathbf{x}) = [[ T(\mathbf{x})=t]]\binom{3}{t}^{-1}$, and $[[ \cdot ]]$ is the indicator function.
Hence, the distribution can be parameterized as a unique mixture of four urn processes, where $T$'s value is the number of black balls. Figure~\ref{fig:urn} illustrates the mixture model. The generative process is as follows. First, choose one of the four urns according to the mixing weights $w_t$; then draw three consecutive balls from the chosen urn without replacement.
\end{example}


\section{Exchangeable Variable Models}

We propose exchangeable variable models (EVMs) as a novel family of tractable probabilistic models for classification and probability estimation. While probabilistic graphical models are built on the notion of (conditional) independence and its graphical representation, EVMs are built on the notion of finite (partial) exchangeability. EVMs can model both \emph{negative} and \emph{positive} correlations in what would be high tree-width graphical models without losing tractability of probabilistic inference. 

The basic components of EVMs are tuples $(\mathbf{X}, T)$ where $\mathbf{X}$ is a sequence of \emph{discrete} random variables partially exchangeable with respect to the statistic $T$ with values $\mathcal{T}$.


\subsection{Probabilistic Inference}

We can relate finite partial exchangeability to tractable probabilistic inference (see also \cite{Niepert:2014}). We assume that for every joint assignment $\mathbf{x}$, $P(\mathbf{x})$ can be computed in time $\mathbf{poly}(|\mathbf{X}|)$.

\begin{prop}
  \label{prop-eff}
  Let $\mathbf{X}$ be partially exchangeable with respect to the statistic $T$ with values $\mathcal{T}$, let $|\mathcal{T}| = \mathbf{poly}(|\mathbf{X}|)$, and let, for any partial assignment $\mathbf{e}$, $ S_{t,\mathbf{e}} := \left\{\mathbf{x} \mid T(\mathbf{x}) = t \text{ and } \mathbf{x} \sim \mathbf{e} \right\},$ where $\mathbf{x} \sim \mathbf{e}$ denotes that $\mathbf{x}$ and $\mathbf{e}$ agree on the variables in their intersection~\cite{Koller:2009}.
  If we can in time $\mathbf{poly}(|\mathbf{X}|)$,
  \begin{enumerate}
  \vspace{-2mm}
    \item[(1)] for every $\mathbf{e}$ and every $t \in \mathcal{T}$, decide if there exists an $\mathbf{x} \in S_{t,\mathbf{e}}$ and, if so, construct such an $\mathbf{x}$, 
  \end{enumerate}
  then the complexity of MAP inference, that is, computing $\argmax_{\mathbf{y}} P(\mathbf{y},\mathbf{e})$ for any partial assignment $\mathbf{e}$, is $\mathbf{poly}(|\mathbf{X}|)$.
  If, in addition, we can in time $\mathbf{poly}(|\mathbf{X}|)$,
  \begin{enumerate}
    \vspace{-2mm}
 	\item[(2)] for every $\mathbf{e}$ and every $t \in \mathcal{T}$, compute $|S_{t,\mathbf{e}}|$,
 \end{enumerate}
 then the complexity of marginal inference, that is, computing $P(\mathbf{e})$ for any partial assignment $\mathbf{e}$, is $\mathbf{poly}(|\mathbf{X}|)$.
  \end{prop}

Proposition~\ref{prop-eff} generalizes to probabilistic models where $P(\mathbf{x})$ can only be computed up to a constant factor $Z$ such as undirected graphical models. Please note that computing conditional probabilities is tractable whenever conditions (1) and (2) are satisfied. We say a statistic is tractable if either of the conditions is fulfilled. 

Proposition~\ref{prop-eff} provides a theoretical framework for developing tractable non-local potentials.
For instance, for $n$ exchangeable Bernoulli variables, the complexity of MAP and marginal inference is polynomial in $n$. This follows from the statistic $T$ satisfying conditions (1) and (2) and since $|\mathcal{T}|=n+1$.
Related work on cardinality-based potentials has mostly focused on MAP inference~\cite{Gupta:2007,tarlow2010hop}. Finite exchangeability also speaks to marginal inference via the tractability of computing $U_t(\mathbf{e}) = |S_{t,\mathbf{e}}|^{-1}$. EVMs can model unary potentials using singleton sets of exchangeable variables. 
While not all instances of finite partial exchangeability result in tractable probabilistic models there exist several examples satisfying conditions (1) and (2) which go beyond finite exchangeability. In the supplementary material, in addition to the proofs of all theorems and propositions, we present examples of tractable statistics that are different from those associated with cardinality-based potentials~\cite{Gupta:2007,tarlow2010hop,tarlow2012fast,Bui:2012}.

\subsection{Parameter Learning}

The parameters of finite sequences of partially exchangeable variables are the mixture weights of the parameterization given in Equation~\ref{equation-paramterization} of Theorem~\ref{theorem-param-pe}. Estimating the parameters of these basic components of EVMs is a crucial task. We derive the maximum-likelihood estimates for these mixture weight vectors.

\begin{theorem}
\label{theorem-mle}
Let $X_1, ...,X_n$ be a sequence of random variables with joint distribution $P$, let $T$ be a statistic with distinct values $t_0,...,t_k$, and let $X_1,...,X_n$ be partially exchangeable with respect to $T$. The ML estimates for $N$ examples, $\mathbf{x}^{(1)}, ..., \mathbf{x}^{(N)}$, are $\mathtt{MLE}[(w_0,...,w_k)] = \left(\frac{c_0}{N}, ..., \frac{c_k}{N}\right)$, where $c_i = \sum_{j=1}^{N} [[T\left(\mathbf{x}^{(j)}\right)=t_i]]$.
\end{theorem}

Hence, the statistical parameters to be estimated are identical to the statistical parameters of a multinomial distribution with $|\mathcal{T}|$ distinct categories.

\subsection{Structure Learning}

Let $\mathbf{\hat{X}}$ be a sequence of random variables and let $\mathbf{\hat{x}}^{(1)}, ..., \mathbf{\hat{x}}^{(N)}$ be $N$ iid examples drawn from the data-generating distribution.
In order to learn the structure of EVMs we need to address two problems. 

\textbf{Problem 1:} Find subsequences $\mathbf{X} \subseteq \mathbf{\hat{X}}$ that are exchangeable with respect to a given tractable statistic $T$. This identifies individual EVM components $(\mathbf{X}, T)$ for which tractable inference and learning is possible. We may utilize different tractable statistics for different components. 

\textbf{Problem 2:} Construct graphical models whose potentials are the previously learned tractable EVM components. In order to preserve tractability of the global model, we have to restrict the class of possible graphical structures.

We now present approaches to these two problems that learn expressive EVMs while maintaining tractability.


Let us first address \textbf{Problem 1}. We focus on EVMs with finitely exchangeable components. Fortunately, there exist several \emph{necessary} conditions for  finite exchangeability (see Definition~\ref{full-exch}) of a sequence of random variables. 

\begin{proposition}
\label{prop-criteria}
The following statements are necessary conditions for exchangeability of a finite sequence of random variables $X_1,...,X_n$. For all $i,j,i',j' \in \{1,...,n\}$ with $i\neq j$ and $i' \neq j'$ 
\begin{enumerate}
\item[(1)] $\mathtt{\mathbf{E}}(X_i)=\mathtt{\mathbf{E}}(X_j)$; 
\item[(2)] $\mathtt{\mathbf{Var}}(X_i)=\mathtt{\mathbf{Var}}(X_j)$; and
\item[(3)] $\mathtt{\mathbf{Cov}}(X_i,X_j) = \mathtt{\mathbf{Cov}}(X_{i'},X_{j'})  \geq - \frac{\mathtt{\mathbf{Var}}(X_i)}{(n-1)}$.
\end{enumerate}
\end{proposition}

The necessary conditions can be exploited to assess whether a sequence of variables is finitely exchangeable. In order to learn EVM components $(\mathbf{X}, T)$ we assume that a sequence of variables is exchangeable unless a statistical test contradicts some or all of the necessary conditions for finite exchangeability. For instance, if a statistical test deemed the expectations $\mathtt{\mathbf{E}}(X)$ and $\mathtt{\mathbf{E}}(X')$ for two variables $X$ and $X'$ identical, we could assume $X$ and $X'$ to be exchangeable. If we wanted the statistical test for finite exchangeability to be more specific and less sensitive, we would also require conditions (2) and/or (3) to hold. Please note the analogy to structure learning with conditional independence tests. Instead of  identifying (conditional) independencies we identify finite exchangeability among random variables. For a sequence of identically distributed variables the assumption of exchangeability is \emph{weaker} than that of independence. 
Testing whether two discrete variables have identical mean and variance is efficient algorithmically. Of course, the application of the necessary conditions for finite exchangeability is only one possible approach to learning the components of EVMs. 

Let us now turn to \textbf{Problem 2}. To ensure tractability, the global graphical structure has to be restricted to tractable classes such as chains and trees. Here, we focus on mixture models where, conditioned on the values of the latent variable, $\mathbf{\hat{X}}$ is partitioned into exchangeable blocks (see Figure~\ref{fig:spectrum}). Hence, for each value $y$ of the latent variable, we perform the statistical tests of \textbf{Problem 1} with estimates of the conditional expectations $\mathtt{\mathbf{E}}(X \mid y)$. We introduce this class of EVMs in the next section and leave more complex structures to future work.

In the context of longitudinal studies and repeated-measures experiments, where an observation is made at different times and under different conditions, there exist several models taking into account the correlation between these observations and assuming identical or similar covariance structure for subsets of the variables~\cite{Jennrich:1986}. These compound symmetry models, however, do not make the assumption of exchangeability and, therefore, do not generally facilitate tractable inference. Nevertheless, finite exchangeability can be seen as a form of parameter tying, a method that has also been applied in the context of hidden Markov models, neural networks~\cite{Rumelhart:1986} and, most notably,  statistical relational learning~\cite{Getoor:2007}. Collective graphical models~\cite{Sheldon:2011} (CGMs) and high-order potentials~\cite{tarlow2010hop,tarlow2012fast} (HOPs) are models based on non-local potentials. Proposition~\ref{prop-criteria} can be applied for learning the structure of novel tractable instances of CGMs and HOPs.

\section{Exchangeable Variable Models for Classification and Probability Estimation}

We are now in the position to design model families that combine the notions of (partial) exchangeability with that of (conditional) independence. Instead of specifying a structure that solely models the (conditional) independence characteristics of the probabilistic model, EVMs also specify sequences of variables that are (partially) exchangeable. The previous results provide the necessary tools to learn both the structure and parameters of partially exchangeable sequences and to perform tractable probabilistic inference. 

\begin{figure}[t!]
\begin{center}
\includegraphics[width=0.48\textwidth]{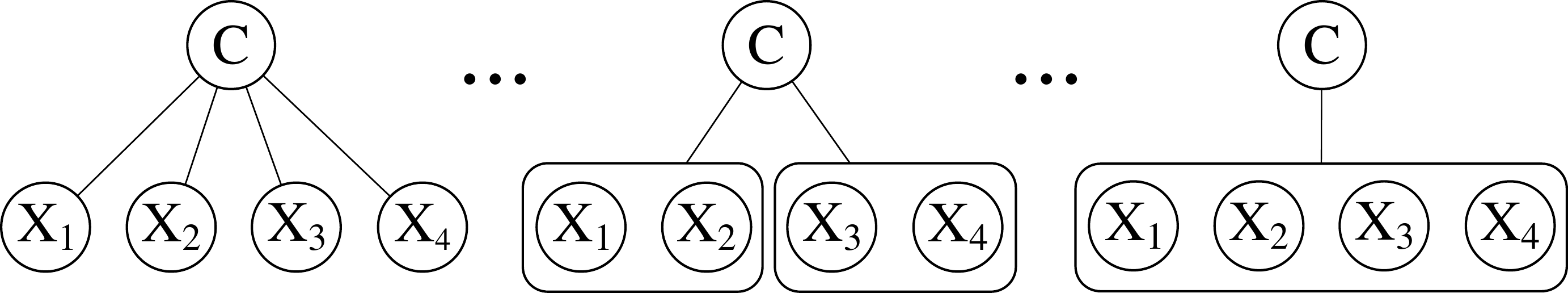}
\caption{\label{fig:spectrum} The combination of exchangeable and independent variables leads to a spectrum of models. On the one end  is the model where, conditioned on the class, all variables are independent (but possibly not identically distributed; left). On the other end is the model where, conditioned on the class, all variables are exchangeable (but possibly correlated; right). The partition of the variables into exchangeable blocks can vary with the class value.}
\end{center}
\end{figure}

The possibilities for building families of exchangeable variable models (EVMs) are vast. Here, we focus on a family of mixtures of EVMs generalizing the widely used naive Bayes model. The family of probabilistic models is therefore also related to research on extending the naive Bayes classifier~\cite{domingos:1997,rennie:2003}. 
The motivation behind this novel class of EVMs is that it facilitates \emph{both} tractable maximum-likelihood learning \emph{and} tractable probabilistic inference.

In line with existing work on mixture models, we derive the maximum-likelihood estimates for the fully observed setting, that is, when there are no examples with missing class labels. We also discuss the expectation maximization (EM) algorithm for the case where the data is partially observed, that is, when examples with missing class labels exist.

\begin{definition}[Mixture of EVMs]
The mixture of EVMs (MEVM) model consists of a class variable $Y$ with $k$ possible values, a set of binary attributes $\mathbf{\hat{X}}=\{X_1, ..., X_n\}$ and, for each $y \in \{1, ..., k\}$, a set $\mathcal{X}_y$ specifying a partition of the attributes into blocks of exchangeable sequences. The structure of the model, therefore, is defined by $\mathcal{X} = \{\mathcal{X}_i\}_{i=1}^{k}$, the set of attribute partitions, one for each class value. The model has the following parameters:

\begin{enumerate}
\item A parameter $p(y)$ for every $y \in \{1, ..., k\}$ specifying the prior probability
of seeing class value $y$. 
\item A parameter $q_{(\mathbf{X})}(\ell \mid y)$ for every $y \in \{1, ..., k\}$, every $\mathbf{X} \in \mathcal{X}_y$, and every $\ell \in \{0, 1, ..., |\mathbf{X}|\}$. The value of $q_{(\mathbf{X})}(\ell \mid y)$ is the probability of the exchangeable sequence $\mathbf{X} \subseteq \mathbf{\hat{X}}$ having an assignment with $\ell$ number of $1$s, conditioned on the class label being~$y$. 
\end{enumerate}
Let $\mathtt{n}_{\mathbf{X}}(\mathbf{\hat{x}})$ be the number of $1$s in the joint assignment $\mathbf{\hat{x}}$ projected onto the variable sequence $\mathbf{X} \subseteq \mathbf{\hat{X}}$. The probability for every $y, \mathbf{\hat{x}}= (x_1, ..., x_n)$ is then defined as $$\mathtt{\textbf{P}}(y, \mathbf{\hat{x}}) = p(y)\prod_{\mathbf{X} \in \mathcal{X}_y} q_{(\mathbf{X})}(\mathtt{n}_{\mathbf{X}}(\mathbf{\hat{x}}) \mid y)\binom{|\mathbf{X}|}{\mathtt{n}_{\mathbf{X}}(\mathbf{\hat{x}})}^{-1}.$$
\end{definition}

Hence, conditioned on the class, the attributes are partitioned into mutually independent and disjoint blocks of exchangeable sequences. 
Figure~\ref{fig:spectrum} illustrates the model family with the naive Bayes model being positioned on one end of the spectrum. Here, $\mathcal{X}_y = \{\{X_1\},...,\{X_n\}\}$ for all $y \in \{1, ..., k\}$. On the other end of the spectrum is the model that assumes full exchangeability conditioned on the class. Here, $\mathcal{X}_y = \{\{X_1,...,X_n\}\}$ for all $y \in \{1, ..., k\}$. For binary attributes, the number of \emph{free} parameters is $k + kn - 1$ for \emph{each} member of the MEVM family. The following theorem provides the maximum-likelihood estimates for these parameters.
 
\begin{theorem}
\label{theorem-em-mevm}
The maximum-likelihood estimates for a MEVM with attributes $\mathbf{\hat{X}}$, structure $\mathcal{X} = \{\mathcal{X}_i\}_{i=1}^{k}$, and a sequence of examples $\left( y^{(i)},\mathbf{\hat{x}}^{(i)}\right), 1 \leq i \leq N,$ are
$$p(y) = \frac{\sum_{i=1}^N [[y^{(i)} = y]]}{N}$$ and, for each $y$ and each $\mathbf{X} \in \mathcal{X}_y$,
$$q_{(\mathbf{X})}(\ell \mid y) = \frac{\sum_{i=1}^{N}[[y^{(i)} = y \mbox{ and } \mathtt{n}_{\mathbf{X}}\hspace{-1mm}\left(\mathbf{\hat{x}}^{(i)}\right) = \ell]]}{\sum_{i=1}^{N}[[y^{(i)}=y]]}.$$
\end{theorem}

We utilize MEVMs for classification problems by learning the parameters and computing the MAP state of the class variable conditioned on assignments to the attribute variables. For probability estimation the class is latent and we can apply  Algorithm~\ref{algorithm-em}. 
The expectation maximization (EM) algorithm is initialized by assigning random examples to the mixture components. In each EM iteration, the examples are fractionally assigned to the components, and the block structure and parameters are updated. Finally, either the previous or current structure is chosen based on the maximum likelihood. For the structure learning step we can, for instance, apply conditions from Proposition~\ref{prop-criteria} where we use the conditional expectations $\mathtt{\mathbf{E}}(X_j \mid y)$, estimated by $\sum_{i=1}^{N}x_j^{(i)}\delta(y \mid i)/N$, for the statistical tests to construct $\mathcal{X}_y$. Since the new structure is chosen from a set containing the structure from the previous EM iteration, the convergence of Algorithm~\ref{algorithm-em} follows from that of structural expectation maximization~\cite{Friedman:1998}. 

\begin{algorithm}[t]
\floatname{require}{Initialize}
\caption{Expectation Maximization for MEVMs}
\label{algorithm-em}
\begin{algorithmic}
\STATE \textbf{Input:} The number of classes $k$. Training examples $\langle \mathbf{\hat{x}}^{(i)}=(x_1^{(i)},...,x_n^{(i)})\rangle, 1 \leq i \leq N$. A parameter specifying a stopping criterion.
\STATE \textbf{Initialization:} Assign $\lfloor N/k\rfloor$ random examples to each mixture component. For each class value $y \in \{1,...,k\}$, partition the $n$ variables into exchangeable sequences $\mathcal{X}^{(0)}_y$, and compute $p^{(0)}(y)$ and $q^{(0)}_{(\mathbf{X})}(\ell \mid y)$ for each $\mathbf{X} \in \mathcal{X}^{(0)}_y$ and $0 \leq \ell \leq |\mathbf{X}|$ using Theorem~\ref{theorem-em-mevm}.
\STATE \textbf{Iterate:} \ \ until stopping criterion is met
\STATE \ \  1. For $i=1,...,N$ and $y = 1,...,k$ compute
\STATE $$\delta(y \mid i) = \frac{\mathtt{\textbf{P}}^{(t-1)}\hspace{-1mm}\left(y,\mathbf{\hat{x}}^{(i)}\right)}{\sum_{j=1}^{k}\mathtt{\textbf{P}}^{(t-1)}\hspace{-1mm}\left(j,\mathbf{\hat{x}}^{(i)}\right)}.$$
\STATE \ \  2. For each $y \in \{1, ..., k\}$, partition the variables  
\STATE \ \ \ \ \ \ into blocks of exchangeable sequences $\mathcal{X}^{(t)}_y$.
\STATE \ \ 3. Update parameters for both $\mathcal{X}_y^{(t-1)}$ and $\mathcal{X}_y^{(t)}$:
\STATE $$p^{(t)}(y) = \frac{\sum_{i=1}^{N} \delta(y \mid i)}{N},$$
$$q^{(t)}_{(\mathbf{X})}(\ell \mid y) = \frac{\sum_{i=1}^{N}[[\mathtt{n}_{\mathbf{X}}\hspace{-1mm}\left(\mathbf{\hat{x}}^{(i)}\right) = \ell]] \ \delta(y \mid i)}{\sum_{i=1}^{N} \delta(y \mid i)}.$$
\STATE \ \ 4. Select the new block structure according to the
\STATE \ \ \ \ \ \ maximum log-likelihood on training examples.
\STATE \textbf{Output:} Structure and parameter estimates.
\end{algorithmic}
\end{algorithm}

A crucial question is how \emph{expressive} the novel model family is. We provide an analytic answer by showing that MEVMs are globally optimal under zero-one loss for a large class of Boolean functions, namely, conjunctions and disjunctions of attributes and symmetric Boolean functions. \emph{Symmetric Boolean functions} are Boolean function whose value depends only on the number of ones in the input~\cite{canteaut:2005}. The class includes (a) Threshold functions, whose value is $1$ on inputs vectors with $k$ or more ones for a fixed $k$; (b) Exact-value functions, whose value is $1$ on inputs vectors with $k$ ones for a fixed $k$; (c) Counting functions, whose value is $1$ on inputs vectors with the number of ones congruent to $k\ \mathtt{mod}\ m$ for fixed $k, m$; and (d) Parity functions, whose value is $1$ if the input vector has an odd number of ones. 

\begin{definition}\cite{domingos:1997}
The Bayes rate for an example is the lowest zero-one loss achievable by any classifier on that example. A classifier is \emph{locally optimal} for an example iff its zero-one loss on that example is equal to the Bayes rate. A classifier is \emph{globally optimal} for a sample iff it is locally optimal for every example in that sample. A classifier is globally optimal for a problem iff it is globally optimal for all possible samples of that problem.
\end{definition}


We can now state the following theorem.

\begin{theorem}
\label{optimality}
The mixtures of EVMs family is globally optimal under zero-one loss for 
\begin{enumerate}
\item Conjunctions and disjunctions of attributes;
\item Symmetric Boolean functions such as
\begin{itemize}
\item Threshold (m-of-n) functions 
\item Parity functions
\item Counting functions
\item Exact value functions
\end{itemize}
\end{enumerate}
\end{theorem}

Theorem~\ref{optimality} is striking as the parity function and its special case, the XOR function, are instances of not linearly separable functions which are often used as examples of particularly challenging classification problems. 
The optimality for symmetric Boolean functions holds even for the model that  assumes \emph{full} exchangeability of the attributes given the value of the class variable (see Figure~\ref{fig:spectrum}, right). It is known that the naive Bayes classifier is \emph{not} globally optimal for threshold (m-of-n) functions despite them being linearly separable~\cite{domingos:1997}. Hence, combining conditional independence and exchangeability leads to highly tractable probabilistic models that are globally optimal for a broader class of Boolean functions. 

\section{Experiments} 

We conducted extensive experiments to assess the efficiency and effectiveness of MEVMs as tractable probabilistic models for classification and probability estimation. A major objective is the comparison of MEVMs and naive Bayes models. We also compare MEVMs with several state of the art classification algorithms. For the probability estimation experiments, we compare MEVMs to latent naive Bayes models and several widely used tractable graphical model classes such as latent tree models.

\begin{table}[t!]
\caption{\label{table-property-class} Properties of the classification data sets and mean and standard deviation of the number of MEVM blocks.}
\small
\begin{center}
\begin{tabular}{|l||r|r|r|r|}
\hline 
Data set & $|V|$ & Train & Test & Blocks \\  
\hline 
\hline
Parity & 1,000 & $10^6$ & 10,000 & $1.3 \pm 0.3$ \\
Counting & 1,000 & $10^6$ & 10,000 & $1.9 \pm 0.9$  \\
M-of-n & 1,000 & $10^6$ & 10,000 & $2.4 \pm 1.6$  \\
Exact & 1,000 & $10^6$ & 10,000 & $3.2 \pm 2.1$ \\
\hline
\hline
20Newsgrp & 19,726.1 & 1,131.4 & 753.2 & $19.2 \pm 1.5$  \\ 
Reuters-8 & 19,398.0 & 1,371.3 & 547.2 & $16.9 \pm 9.1$\\ 
Polarity & 38,045.8 & 1,800.0 & 200.0 & $34.1 \pm  0.7$ \\ 
Enron & 43,813.6 & 4,000.0 & 1,000.0 & $30.2 \pm 6.0$  \\
WebKB &  7,290.0 &  1,401.5 & 698.0 & $19.3 \pm  3.6$  \\ 
MNIST & 784.0 & 12,000.0 & 2,000.0 & $72.3 \pm 3.1$\\
\hline
\end{tabular} 
\end{center}
\end{table}

\subsection{Classification}

We evaluated the MEVM classifier using both synthetic and real-world data sets. Each synthetic data set consists of $10^6$ training and $10000$ test examples. Let $\mathtt{n}(\mathbf{x})$ be the number of ones of the example $\mathbf{x}$. The parity data was generated by sampling uniformly at random an example $\mathbf{x}$ from the set $\{0,1\}^{1000}$ and assigning it to the first class if $\mathtt{n}(\mathbf{x})\ \mathtt{ mod }\ 2 = 1$, and to the second class otherwise. For the $10$-of-$1000$ data set we assigned an example $\mathbf{x}$ to the first class if $\mathtt{n}(\mathbf{x}) \geq 10$, and to the second class otherwise. For the counting data set we assigned an examples $\mathbf{x}$ to the first class if $\mathtt{n}(\mathbf{x})\ \mathtt{ mod }\ 5 = 3$, and to the second class otherwise. For the exact data set we assigned an example $\mathbf{x}$ to the first class if $\mathtt{n}(\mathbf{x}) \in \{0,200,400,600,800,1000\}$, and to the second class otherwise.

We used the \textsc{SciKit} $0.14$\footnote{http://scikit-learn.org/} functions to load the 20Newsgroup train and test samples. 
We removed headers, footers, and quotes from the training and test documents. This renders the classification problem more difficult and leads to significantly higher zero-one loss for all classifiers. For the Reuters-8 data set we considered only the Reuters-21578 documents  with a single topic and the top $8$ classes that have at least one train and one test example. For the WebKB text data set we considered the classes $\mathtt{project}$, $\mathtt{course}$, $\mathtt{faculty}$, and $\mathtt{student}$. For all text data sets we used the binary bag-of-word representation resulting in feature spaces with up to $45000$ dimensions. 
For the MNIST data set, a collection of hand-written digits, we set a feature value to $1$ if the original feature value was greater than $50$, and to $0$ otherwise. The polarity data set is a well-known sentiment analysis problem based on movie reviews~\cite{Pang:2004}. The problem is to classify movie reviews as either positive or negative. We used the cross-validation splits provided by the authors. 
The Enron spam data set is a collection of e-mails from the Enron corpus that was divided into spam and no-spam messages~\cite{Metsis:2006}. Here, we applied randomized $100$-fold cross validation. We did not apply feature extraction algorithms to any of the data sets.
Table~\ref{table-property-class} lists the properties of the data sets and the mean and standard deviation of the number of blocks of the MEVMs.
We distinguished between two-class and multi-class (more than $2$ classes) problems. When the original data set had more than two classes, we created the two-class problems by considering every pair of classes as a separate cross-validation problem. We draw this  distinction because we want to compare classification approaches independent of particular multi-class strategies (1-vs-n, 1-vs-1, etc.).

\begin{table}[t!]
\caption{\label{table-results-twoclass} Accuracy values for the two-class experiments. Bold numbers indicate significance (paired t-test; $p < 0.01$) compared to non-bold results in the same row.}
\small
\begin{center}
\begin{tabular}{|l||c|c|c|c|c|}
\hline 
Data set & MEVM & NB & DT & SVM & $5$-NN \\  
\hline 
\hline
Parity & \textbf{0.958} & 0.497 & 0.501 & 0.493 & 0.502 \\
Counting & \textbf{0.967} &  0.580 & 0.655 &  0.768 & 0.765 \\
M-of-n & \textbf{0.994} & 0.852 & 0.990 &  \textbf{0.995} & 0.715 \\
Exact & \textbf{0.996} & 0.566 & 0.983 & \textbf{0.995} &  0.974 \\
\hline
\hline
20Newsgrp & \textbf{0.905} & 0.829 & 0.803 & 0.867 & 0.582 \\ 
Reuters-8 & \textbf{0.968} & 0.940  & \textbf{0.965} & \textbf{0.982} & 0.881 \\ 
Polarity & 0.826 & 0.794 & 0.623 & \textbf{0.859} & 0.520 \\ 
Enron &  \textbf{0.980} & 0.915 & 0.948 & 0.972 &  0.743 \\
WebKB &\textbf{0.943} &  0.907 & 0.899 & \textbf{0.952} & 0.780 \\ 
MNIST & 0.969 & 0.964 & 0.981 & 0.983 & \textbf{0.995}\\
\hline
 \end{tabular} 
\end{center}
\caption{\label{table-results-multiclass} Accuracy values for the multi-class  experiments. Bold numbers indicate significance (paired t-test; $p < 0.01$) compared to non-bold results in the same column.}
 \small
 \begin{center}
 \begin{tabular}{|l||c|c|c|c|}
 \hline 
 Classifier & 20Newsgrp & Reuters-8 & WebKB &  MNIST \\  
 \hline 
  \hline
MEVM & \textbf{0.626} & \textbf{0.911} & \textbf{0.860} &  \textbf{0.855}  \\
NB &  0.537 & 0.862  & 0.783  &  0.842   \\
\hline
 \end{tabular} 
\end{center}
\end{table}

We exploited necessary condition (1) from Proposition~\ref{prop-criteria} to learn the block structure of the MEVM classifiers. For each pair of variables $X, X'$ and each class value $y$, we applied Welch's t-test to test the null hypothesis $\mathtt{\mathbf{E}}(X \mid y) = \mathtt{\mathbf{E}}(X' \mid y)$. If, for two variables, the test's p-value was less than $0.1$, we rejected the null hypothesis and placed them in different blocks conditioned on $y$. We applied Laplace smoothing with a constant of $0.1$. The same parameter values were applied across \emph{all} data sets and experiments.
For all other classifiers we used the \textsc{SciKit} $0.14$ implementations  naive\_bayes.BernoulliNB, tree.DecisionTreeClassifier,  svm.LinearSVC, and neighbors.KNeighborsClassifier. We used the classifiers'  standard  settings except for the naive Bayes classifier where we applied a Laplace smoothing constant (alpha) of $0.1$ to ensure a fair comparison (NB results deteriorated for alpha values of $1.0$ and $0.01$). The standard setting for the classifiers are available as part of the \textsc{SciKit} $0.14$ documentation.
All implementations and data sets will be published.

Table~\ref{table-results-twoclass} lists the results for the two-class problems. The MEVM classifier was one of the best classifiers for $8$ out of the $10$ data sets. With the exception of the MNIST data set, where the difference was insignificant, MEVM significantly outperformed the naive Bayes classifier (NB) on all data sets. The MEVM classifier outperformed SVMs on $4$ data sets, two of which are real-world text classification problems and achieved a tie on $4$. For the parity data set only the MEVM classifier was better than random.
Table~\ref{table-results-multiclass} shows the results on the multi-class problems. Here, the MEVM classifier significantly outperforms naive Bayes on all data set. The MEVM classifier outperformed all classifiers on the 20Newsgroup and was a close second on the Reuters-8 and WebKB  data sets. The MEVM classifier is particularly suitable for high-dimensional and sparse data sets. We hypothesize that this has three reasons. First, MEVMs can model both negative and positive correlations between variables. Second, MEVMs perform a non-linear transformation of the feature space. Third, MEVMs cluster noisy variables into blocks of exchangeable sequences which acts as a form of regularization in sparse domains.

\subsection{Probability Estimation}
 
We conducted experiments with a widely used collection of data sets~\cite{Haaren:2012,Gens:2013,Lowd:2013}.
Table~\ref{table-de-dataset} lists the number of variables, training and test examples, and the number of blocks of the MEVM models. We set the latent variable's domain size to $20$ for each problem and applied the same EM initialization for MEVMs and NB models. This way we could compare NB and MEVM independent of the tuning parameters specific to EM. We implemented EM exactly as described in Algorithm~\ref{algorithm-em}. For step (2), we exploited Proposition~\ref{prop-criteria}~(1) and, for each $y$, partitioned the variables into exchangeable blocks by performing a series of Welch's t-tests on the expectations $\mathtt{\mathbf{E}}(X_j \mid y)$, estimated by $\sum_{i=1}^{N}x_j^{(i)}\delta(y \mid i)/N$, assigning two variables to different blocks if the null hypothesis of identical means could be rejected at a significance level of $0.1$. For MEVM and NB we again used a Laplace smoothing constant of $0.1$. We ran EM until the average log-likelihood increase between iterations was less than $0.001$. We restarted EM $10$ times and chose the model with the maximal log-likelihood on the training examples. We did not use the validation data. For LTM~\cite{Choi:2011}, we applied the four  methods, CLRG, CLNJ, regCLRG, and regCLNJ, and chose the model with the highest validation log-likelihood.

 \begin{table}[t!]
 \caption{\label{table-de-dataset} Properties of the data sets used for probability estimation and mean and standard deviation of the number of MEVM blocks.}
 \small
 \begin{center}
 \begin{tabular}{|l|r|r|r|r|}
 \hline 
 Data set & $|V|$ & Train & Test & Blocks \\  
 \hline 
 NLTCS & 16 & 16,181 &  3,236 & $8.8 \pm 1.9$ \\ 
 MSNBC & 17 & 291,326 &  58,265  & $15.9 \pm 1.1$\\
 KDDCup 2000& 64 & 180,092 &  34,955 &  $15.8 \pm 4.7$\\
 Plants & 69 & 17,412 &  3,482 & $15.9 \pm 2.9$\\ 
 Audio & 100 & 15,000 &  3,000 & $13.7 \pm 3.0$ \\ 
 Jester & 100 & 9,000 &  4,116 & $10.4 \pm 2.0$ \\ 
 Netflix & 100 & 15,000  & 3,000 &  $14.8 \pm 3.2$\\ 
 MSWeb & 294 & 29,441 &  5,000 & $21.3 \pm 2.0$ \\ 
 Book & 500 & 8,700 &  1,739 & $12.4 \pm 2.9$ \\ 
 WebKB & 839 & 2,803 &  838 & $10.6 \pm 2.3$ \\ 
 Reuters-52 & 889 & 6,532 &  1,540 & $16.7 \pm 3.1$ \\ 
 20Newsgroup & 910 & 11,293 &  3,764 & $17.9 \pm 3.7$ \\ 
 \hline
 \end{tabular}
\end{center}
\end{table}

Table~\ref{table-density-results-ll} lists the average log-likelihood of the test data for the MEVM, the latent naive Bayes~\cite{Lowd:2005} (NB), the latent tree (LTM), and the Chow-Liu tree model~\cite{Chow:2006} (CL). Even without exploiting the validation data for model tuning, the MEVM models outperformed the CL models on all, and the LTMs on all but two of the data set. MEVMs achieve the highest log-likelihood score on $7$ of the $12$ data sets. With the exception of the Jester data set, MEVMs either outperformed or tied the NB model. While the results indicate that MEVMs are effective for higher-dimensional and sparse data sets, where the increase in log-likelihood was most significant, MEVMs also outperformed the NB models on $3$ data sets with less than $100$ variables. The MEVM and NB models have exactly the same number of free parameters. 
Since results on the same data sets are available for other tractable model classes we also  compared MEVMs with SPNs~\cite{Gens:2013} and ACMNs~\cite{Lowd:2013}.
Here, MEVMs are outperformed by the more complex SPNs on $5$ and by ACMNs on $6$ data sets. However, MEVMs are competitive and outperform SPNs on $7$ and ACMNs on $6$ of the $12$ data sets. Following previous work~\cite{Haaren:2012}, we applied the Wilcoxon signed-rank test. MEVM outperforms the other models at a significance level of $0.0124$ (NB), $0.0188$ (LTM), and $0.0022$ (CL). The difference is insignificant compared to ACMNs ($0.6384$) and SPNs ($0.7566$). 

To compute the probability of one example, MEVMs require as many steps as there are blocks of exchangeable variables.  Hence, EM for MEVM is significantly more efficient than EM for NB, both for a single EM iteration and to reach the stopping criterion. While the difference was less significant for problems with fewer than $100$ variables, the EM algorithm for MEVM was up to \emph{two orders of magnitude} faster for data sets with $100$ or more variables.  

\section{Discussion}


\begin{table}[t!]
\caption{\label{table-density-results-ll} Average log-likelihood of the MEVM, the naive Bayes, the latent tree, and the Chow-Liu tree model.}
\small
\vspace{-0.38mm}
\begin{center}
\begin{tabular}{|l|r|r|r|r|}
\hline 
Data set & MEVM & NB & LTM & CL \\  
\hline 
NLTCS & -6.04 & -6.04 & -6.46 & -6.76 \\ 
MSNBC & \textbf{-6.23} & -6.71 & -6.52 & -6.54 \\
KDDCup 2000& \textbf{-2.13} & -2.15 & -2.18 & -2.29 \\
Plants & \textbf{-14.86} & -15.10 & -16.39 & -16.52 \\ 
Audio & \textbf{-40.63} & -40.69 & -41.89 & -44.37 \\ 
Jester & -53.22 & \textbf{-53.19} & -55.17 &  -58.23 \\ 
Netflix & \textbf{-57.84} & -57.87 & -58.53 & -60.25 \\ 
MSWeb & -9.96 & -9.96 & -10.21 & -10.19 \\ 
Book & -34.63 & -34.80 & \textbf{-34.23} & -34.70  \\ 
WebKB & -157.21 & -158.01 & \textbf{-156.84} & -163.48  \\ 
Reuters-52 & \textbf{-86.98} & -87.32 & -91.25 & -94.37 \\ 
20Newsgroup & \textbf{-152.69} & -152.78 & -156.77 & -164.13\\ 
\hline
\end{tabular}
\end{center}
\end{table}

Exchangeable variable models (EVMs) provide a framework for probabilistic models combining the notions of conditional independence and partial exchangeability.  As a result, it is possible to efficiently learn the parameters and structure of tractable high tree-width models. EVMs can model complex positive and negative correlations between large numbers of variables. We presented the theory of EVMs and showed that  a particular subfamily is optimal for several important classes of Boolean functions. Experiments with a large number of data sets verified that mixtures of EVMs are powerful and highly efficient models for classification and probability estimation.

EVMs are potential components in deep architectures such as sum-product networks~\cite{Gens:2013}. In light of Theorem~\ref{optimality}, exchangeable variable nodes, complementing sum and product nodes, can lead to more compact representations with fewer parameters to learn.
EVMs  are also related to graphical modeling with perfect graphs~\cite{Jebara:2013}. In addition, EVMs provide an insightful connection to lifted probabilistic inference~\cite{Kersting:2012}, an active research area concerned with exploiting symmetries for more efficient probabilistic inference. We have developed a principled framework based on partial exchangeability as an important notion of structural symmetry. There are numerous opportunities for cross-fertilization between EVMs, perfect graphical models, collective graphical models, and statistical relational models. 

Directions for future work include more sophisticated structure learning, EVMs with continuous variables, EVMs based on instances of partial exchangeability other than finite exchangeability, novel statistical relational formalisms incorporating EVMs, applications of EVMs, and a general theory of graphical models with exchangeable potentials.

\section*{Acknowledgments}
\small
Many thanks to Guy Van den Broeck, Hung Bui, and Daniel Lowd for helpful discussions. This research was partly funded by ARO grant W911NF-08-1-0242,
ONR grants N00014-13-1-0720 and N00014-12-1-0312, and AFRL contract
FA8750-13-2-0019. The views and conclusions contained in this document are those of the authors and should not be interpreted as necessarily representing the official policies, either expressed or implied, of ARO, ONR, AFRL, or the United States Government.

\newpage
\small
\bibliography{evm}
\bibliographystyle{icml2014}
\normalsize

\newpage
\appendix

%
%

\section{Proof of Proposition~\ref{prop-eff}}

Let $\mathbf{X}$ be partially exchangeable with respect to the statistic $T$ with values $\mathcal{T}$, let $|\mathcal{T}| = \mathbf{poly}(|\mathbf{X}|)$, and let, for any partial assignment $\mathbf{e}$, $ S_{t,\mathbf{e}} := \left\{\mathbf{x} \mid T(\mathbf{x}) = t \text{ and } \mathbf{x} \sim \mathbf{e} \right\},$ where $\mathbf{x} \sim \mathbf{e}$ denotes that $\mathbf{x}$ and $\mathbf{e}$ agree on the variables in their intersection~\cite{Koller:2009}.
  If we can in time $\mathbf{poly}(|\mathbf{X}|)$,
  \begin{enumerate}
    \item[(1)] for every $\mathbf{e}$ and every $t \in \mathcal{T}$, decide if there exists an $\mathbf{x} \in S_{t,\mathbf{e}}$ and, if so, construct such an $\mathbf{x}$, 
  \end{enumerate}
  then the complexity of MAP inference, that is, computing $\argmax_{\mathbf{y}} P(\mathbf{y},\mathbf{e})$ for any partial assignment $\mathbf{e}$, is $\mathbf{poly}(|\mathbf{X}|)$.
  If, in addition, we can in time $\mathbf{poly}(|\mathbf{X}|)$,
  \begin{enumerate}
 	\item[(2)] for every $\mathbf{e}$ and every $t \in \mathcal{T}$, compute $|S_{t,\mathbf{e}}|$,
 \end{enumerate}
 then the complexity of marginal inference, that is, computing $P(\mathbf{e})$ for any partial assignment $\mathbf{e}$, is $\mathbf{poly}(|\mathbf{X}|)$.

\begin{proof}
We first prove statement (1). Let $\mathbf{e}$ be a given partial assignment and assume we want to compute $\argmax_{\mathbf{y}} P(\mathbf{y},\mathbf{e})$. We construct an $\mathbf{x}_t \in S_{t,\mathbf{e}}$ for each $t \in \mathcal{T}$ and set $\hat{\mathbf{x}}_t := \argmax_{\mathbf{x}_t} P(\mathbf{x}_t)$. By assumption, this is possible in time $\mathbf{poly}(|\mathbf{X}|)$.  Since we have that  $\hat{\mathbf{x}}_t = \mathbf{\hat{y}e}$ with $\mathbf{\hat{y}} := \argmax_{\mathbf{y}} P(\mathbf{y},\mathbf{e})$ we can extract the solution in linear time.

To prove statement (2), let $\mathbf{e}$ be a partial assignment. We construct a $\mathbf{x}_t \in S_{t,\mathbf{e}}$ for each $t \in \mathcal{T}$ for which such an $\mathbf{x}_t$ exists, compute $|S_{t,\mathbf{e}}|$, and return $\sum_{t \in \mathcal{T}} P(\mathbf{x}_t)|S_{t,\mathbf{e}}|$. By assumption, this is possible in time $\mathbf{poly}(|\mathbf{X}|)$.
\end{proof}

We can utilize Proposition~\ref{prop-eff} to prove that probabilistic inference for a sequence of $n$ exchangeable binary variables is tractable. 

\begin{example}[Finite Exchangeability]
\label{example-simple}
Let $\mathbf{X}$ be an exchangeable sequence of binary random variables. Let $\mathtt{n}(\mathbf{e})$ be the number of $1$s in a partial assignment $\mathbf{e}$ to the variables $\mathbf{X}$. Clearly, we have that $\mathbf{X}$ is exchangeable with respect to the statistic $T(\mathbf{x}) = \mathtt{n}(\mathbf{x})$ with values $\mathcal{T} = \{0,...,n\}$.

First, we prove that for every partial assignment $\mathbf{e}$ to $k$ of the $n$ variables and every $t \in \mathcal{T}$, we can decide if there exists an $\mathbf{x} \in S_{t,\mathbf{e}}$ and, if so, construct such an $\mathbf{x}$ in time $\mathbf{poly}(|\mathbf{X}|)$. If $\mathtt{n}(\mathbf{e}) > t$ or $n - k + \mathtt{n}(\mathbf{e}) < t$, then there does \emph{not} exist such an $\mathbf{x}$. Otherwise it is possible to  generate a $\mathbf{x}$ with $\mathtt{n}(\mathbf{x}) = t$ in linear time by assigning exactly $t - \mathtt{n}(\mathbf{e})$ ones to the unassigned variables and we have that $\mathbf{x} \in S_{t,\mathbf{e}}$. Hence, MAP inference is tractable.

Next, we prove that for every partial assignment $\mathbf{e}$ to $k$ variables and every $t \in \mathcal{T}$, we can compute $|S_{t,\mathbf{e}}|$ in time $\mathbf{poly}(|\mathbf{X}|)$. But this is possible since $|S_{t,\mathbf{e}}| = \binom{n - k}{t - \mathtt{n}(\mathbf{e})}$. Hence, marginal inference is tractable.
\end{example}

Please note that Example~\ref{example-simple} implies tractability results for numerous important special cases of finite exchangeability such as parity and threshold functions.

There are forms of finite partial exchangeability~\cite{Diaconis:1980} that go beyond the notion of \emph{full} finite exchengeability and, therefore, cardinality-based potentials~\cite{Gupta:2007,tarlow2010hop} of Example~\ref{example-simple}. We provide three examples.

\begin{example}[Block Exchangeability]
Let $w$ be a fixed constant. For a sequence of binary random variables $\mathbf{X}$ let $\mathcal{X} = \{\mathbf{X}_1,...,\mathbf{X}_w\}$ be a partition of the variables $\mathbf{X}$ into $w$ subsequences, and let $\mathtt{n}_{\mathbf{Y}}(\mathbf{x})$ be the number of $1$s in an assignment $\mathbf{x}$ projected onto the variables $\mathbf{Y} \subseteq \mathbf{X}$. Now, let $T(\mathbf{x}) = (\mathtt{n}_{\mathbf{X}_1}(\mathbf{x}), ..., \mathtt{n}_{\mathbf{X}_w}(\mathbf{x}))$.

It is straight-forward to verify that $|\mathcal{T}| = \mathbf{poly}(|\mathbf{X}|)$. Moreover, with arguments similar to those made in Example~\ref{example-simple} one can show that conditions (1) and (2) of Proposition~\ref{prop-eff} are met. Hence, MAP and marginal inference are tractable for the statistic $T$.
\end{example}

\begin{example}
Let $\mathbf{X}$ be a sequence of $n$ binary random variables and let  $\tau_{0\rightarrow 1}(\mathbf{x})$ be the number of times $01$ occurs as a substring\footnote{As opposed to subsequences, substrings are consecutive parts of a string.} in $\mathbf{x}$. Now, consider the  statistic
$$T(\mathbf{x}) = \tau_{0\rightarrow 1}(\mathbf{x}).$$
For example, for $\mathbf{x} = 11011111$ we have $T(\mathbf{x}) = 1$ and for $\mathbf{x} = 01010101$ we have $T(\mathbf{x}) = 4$. We also have that $|\mathcal{T}| = \lfloor n / 2 \rfloor + 1 = \mathbf{poly}(|\mathbf{X}|)$.

Now, let $\mathbf{e}$ be a partial assignment to $k$ of the $n$ variables and let $0 \leq t \leq  \lfloor n / 2 \rfloor$ be a value of the statistic. 
Let $\mathbf{b} = \{0,1,*\}^n$ be a string where the characters $0$ and $1$ encode the assignments to variables according to $\mathbf{e}$ and the character * encodes unassigned variables. We now partition $\mathbf{b}$ into four sets $G_{ij}$, $i,j \in \{0,1\}$, of substrings defined as $G_{ij} := \{ \mathbf{s} \sqsubseteq \mathbf{b} \mid s_1 = i, s_{|\mathbf{s}|}=j, s_{\ell} = \mbox{*} \mbox{ for } 1 \leq i < \ell < j \leq |\mathbf{s}|\},$ where $\sqsubseteq$ denotes the substring relation. We can now complete the partial assignment $\mathbf{e}$ to a joint assignment $\mathbf{x}$ with $T(\mathbf{x}) = t$ if and only if (1) $\tau_{0\rightarrow 1}(\mathbf{b}) + |G_{01}| \leq t$ and (2) $\tau_{0\rightarrow 1}(\mathbf{b}) + \sum_{\mathbf{s} \in G_{00}} \left\lceil \frac{|\mathbf{s}|-2}{2} \right\rceil +
 \sum_{\mathbf{s} \in G_{01}} \left\lfloor \frac{|\mathbf{s}|}{2} \right\rfloor +
  \sum_{\mathbf{s} \in G_{10}} \left\lfloor \frac{|\mathbf{s}|-2}{2} \right\rfloor +
  \sum_{\mathbf{s} \in G_{11}} \left\lceil \frac{|\mathbf{s}|-2}{2} \right\rceil \geq t$. When these two conditions are met, the full assignment $\mathbf{x}$ can be constructed by completing the substring in the groups $G_{ij}$ so as to make $T(\mathbf{x}) = t$ and this is possible in linear time. Hence, MAP inference is tractable.
\end{example}

It is possible to construct novel tractable statistics by nesting  statistics that are known to be tractable.

\begin{example}[Nested Tractable Statistics]
Let $\mathbf{X}$ be an $n \times n$ array of binary random variables. For instance, $\mathbf{X}$ could represent a binarized image with $n$ rows and $n$ columns. Let $k$ be a fixed integer constant and let $\ell$ be the integer such that $n = k \ell$. We assume without loss of generality that such an integer exists. We partition the original array into $\ell^2$ squares of dimension $k \times k$. For $1 \leq i \leq \ell^2$, let $\mathbf{S}_i$ be the variables of square $i$. Now, let $T_1: \{0,1\}^{k^2} \rightarrow \{0,1\}$ be the statistic defined as $$T_1(\mathbf{s}=(s_1,...,s_{k^2})) = [[\sum_{i = 1}^{k^2}s_i > \tau]],$$
for some $\tau$ with $0 \leq \tau < k^2$. That is, $T_1(\mathbf{s}) = 1$, if the number of $1$s in a given square exceeds a threshold of $\tau$ and $T_1(\mathbf{s}) = 0$ otherwise. Please note that for $\tau=0$ this corresponds to max-pooling. Now, let $T: \{0,1\}^{n^2} \rightarrow \{0,...,\ell^2\}$ be the statistic defined as follows:
$$T(\mathbf{x}) = \sum_{i=1}^{\ell^2} T_1(\mathbf{s}_i).$$
Based on the tractability of the two statistics, it is straight-forward to verify that both MAP and marginal inference is tractable for the statistic $T$.
\end{example}

Please note that the presented theoretical framework facilitates the discovery and development of novel tractable non-local potentials.

\section{Proof of Theorem~\ref{theorem-mle}}

Let $X_1, ...,X_n$ be a sequence of random variables with joint distribution $P$, let $T$ be a statistic with distinct values $t_0,...,t_k$, and let $X_1,...,X_n$ be partially exchangeable with respect to $T$. The ML estimates for $N$ examples, $\mathbf{x}^{(1)}, ..., \mathbf{x}^{(N)}$, are $\mathtt{MLE}[(w_0,...,w_k)] = \left(\frac{c_0}{N}, ..., \frac{c_k}{N}\right)$, where $c_i = \sum_{j=1}^{N} [[T\left(\mathbf{x}^{(j)}\right)=t_i]]$.

\begin{proof}
Let $\theta = (w_0,...,w_k)$. By Theorem~\ref{theorem-param-pe}, the log-likelihood for $N$ examples $\mathbf{x}^{(1)},...,\mathbf{x}^{(N)}$ is
$$\mathtt{L}(\theta) =  \sum_{j=1}^{N}\log\left(\sum_{i=0}^{k} w_i U_i\left(\mathbf{x}^{(j)}\right)\right).$$ 
Let $c_i = \sum_{j=1}^{N} [[T\left(\mathbf{x}^{(j)}\right)=t_i]]$ and let $\hat{\mathbf{x}}_i$ be a joint assignment with $T(\mathbf{\hat{x}}_i)=t_i$. Then, $\mathtt{L}(\theta) = \sum_{i=0}^{k} c_i \log(w_i U_i(\mathbf{\hat{x}}_i)) = \sum_{i=0}^{k} c_i [\log(w_i) + \log(U_i(\mathbf{\hat{x}}_i))] = \sum_{i=0}^{k} c_i \log(w_i) + \sum_{i=0}^{k} c_i \log(U_i(\mathbf{\hat{x}}_i)).$ The second term is free of parameters and, hence, finding the ML estimates amounts to maximizing the first sum. This is equivalent to finding the maximum likelihood estimate of a multinomial which can be solved with Lagrange multipliers. Hence, $\mathtt{MLE}(w_i) = \frac{c_i}{N}$, for $0 \leq i \leq k$. 
\end{proof}

\section{Proof of Proposition~\ref{prop-criteria}}

The following statements are necessary conditions for exchangeability of a finite sequence of random variables $X_1,...,X_n$. For all $i,j,i',j' \in \{1,...,n\}$ with $i\neq j$ and $i' \neq j'$ 
\begin{enumerate}
\item[(1)] $\mathtt{\mathbf{E}}(X_i)=\mathtt{\mathbf{E}}(X_j)$; 
\item[(2)] $\mathtt{\mathbf{Var}}(X_i)=\mathtt{\mathbf{Var}}(X_j)$; and
\item[(3)] $\mathtt{\mathbf{Cov}}(X_i,X_j) = \mathtt{\mathbf{Cov}}(X_{i'},X_{j'})  \geq - \frac{\mathtt{\mathbf{Var}}(X_i)}{(n-1)}$.
\end{enumerate}

These conditions are well-known and are straight-forward to prove. Nevertheless, for the sake of completeness, we prove statement (3).

\begin{proof}
It is straight-forward to prove statements (1) and (2). In order to prove statement (3) we use statements (2) to write
\begin{align}
0 & \le \mathbf{Var}(X_1 + \cdots + X_n) \notag\\
& = \mathbf{Var}(X_1) + \cdots + \mathbf{Var}(X_n) + 2 \sum_{i<j}\mathbf{Cov}(X_i,X_j) \notag\\
& = n\mathbf{Var}(X_i) + n(n-1) \mathbf{Cov}(X_i,X_j). \notag
\end{align}
Hence, $\mathbf{Cov}(X_i,X_j) \geq - \frac{\mathbf{Var}(X_i)}{(n-1)}.$
\end{proof}

\section{Proof of Theorem~\ref{optimality}}

The mixtures of EVMs family is globally optimal under zero-one loss for 
\begin{enumerate}
\item Conjunctions and disjunctions of attributes;
\item Symmetric Boolean functions such as
\begin{itemize}
\item Threshold (m-of-n) functions 
\item Parity functions
\item Counting functions
\item Exact value functions
\end{itemize}
\end{enumerate}

\begin{proof}
Let $\mathbf{X}$ be the sequence of variables under consideration. We write $y(\mathbf{x})$ for the (hidden) class value of example $\mathbf{x}$. For conjunctions of attributes, let $\mathbf{\hat{X}} \subseteq \mathbf{X}$ be the sequence of variables that are part of the conjunction. Conditioned on the binary class variable being either $0$ or $1$, we partition the variables into the two blocks $\mathbf{\hat{X}}$ and $\mathbf{X} - \mathbf{\hat{X}}$. 
We set the parameters of the MEVM as follows.

$q_{(\mathbf{\hat{X}})}(\ell \mid 1) = 1.0$ if $\ell=|\mathbf{\hat{X}}|$ and $q_{(\mathbf{\hat{X}})}(\ell \mid 1) = 0.0$ otherwise;

$q_{(\mathbf{\hat{X}})}(\ell \mid 0) = 0.0$ if $\ell=|\mathbf{\hat{X}}|$ and $q_{(\mathbf{\hat{X}})}(\ell \mid 0) = \frac{\binom{|\mathbf{\hat{X}}|}{\ell}}{2^{|\mathbf{\hat{X}}|}}$ otherwise;

$q_{(\mathbf{X-\hat{X}})}(\ell \mid 1) = \frac{\binom{|\mathbf{X|-|\hat{X}}|}{\ell}}{2^{|\mathbf{X|-|\hat{X}}|}}$;  \ \ \ $q_{(\mathbf{X-\hat{X}})}(\ell \mid 0) = \frac{\binom{\mathbf{|X|-|\hat{X}|}}{\ell}}{2^{\mathbf{|X|-|\hat{X}|}}}$;

$p(1) = \frac{2^{\mathbf{|X|-|\hat{X}|}}}{2^{|\mathbf{X}|}}$; and $p(0) = \frac{(2^{|\mathbf{\hat{X}}|}-1)(2^{\mathbf{|X|-|\hat{X}|}})}{2^{|\mathbf{X}|}}$.

Then, we have that $\mathbf{P}(1 \mid \mathbf{x}) > 0$ if $y(\mathbf{x}) = 1$ and $\mathbf{P}(1 \mid \mathbf{x}) = 0$ otherwise. Moreover,  $\mathbf{P}(0 \mid \mathbf{x}) = 0$ if $y(\mathbf{x}) = 1$ and $\mathbf{P}(0 \mid \mathbf{x}) > 0$ otherwise. Hence, the MEVM classifier always returns the correct class value. A similar argument can be made to prove the optimality for disjunctions of attributes.
 
To prove the second statement, we consider an MEVM model with a binary class variable and the following block structure. For each of the class variable's values $y$, $y \in \{0, 1\}$, we have that $\mathcal{X}_y = \{X_1,...,X_n\}$. That is, conditioned on each class value, the attributes are assumed to be exchangeable (see Figure~\ref{fig:spectrum}; right). It is straightforward to verify that this particular MEVM can learn \emph{arbitrary} discrete distributions over any symmetric Boolean function. 
\end{proof}

\end{document}